\documentclass[11pt,a4paper]{article}
\usepackage[hyperref]{emnlp-ijcnlp-2019}
\usepackage{times}
\usepackage{latexsym}
\usepackage{amsmath}
\usepackage{amssymb}
\usepackage{dblfloatfix}
\usepackage{framed}
\usepackage{graphicx}

\usepackage{url}

\aclfinalcopy % Uncomment this line for the final submission

\title{Repurposing Decoder-Transformer Language Models for Abstractive Summarization}

\author{Luke de Oliveira  \\
  Twilio AI \\ %San Francisco, CA\\
  {\tt ldeoliveira@twilio.com} \\\And
  Alfredo L\'{a}inez Rodrigo \\
  Twilio AI \\ %Madrid, Spain\\
  {\tt alainez@twilio.com} \\}
  
\date{}

\begin{document}
\maketitle
\begin{abstract}

Neural network models have shown excellent fluency and performance when applied to abstractive summarization. Many approaches to neural abstractive summarization involve the introduction of significant inductive bias, exemplified through the use of components such as pointer-generator architectures, coverage, and partially extractive procedures, designed to mimic the process by which humans summarize documents. We show that it is possible to attain competitive performance by instead directly viewing summarization as a language modeling problem and effectively leveraging transfer learning. We introduce a simple procedure built upon decoder-transformers to obtain highly competitive ROUGE scores for summarization performance using a language modeling loss alone, with no beam-search or other decoding-time optimization, and instead relying on efficient nucleus sampling and greedy decoding. 

%We investigate whether competitive performance can be attained without these approaches, by directly viewing summarization as a language modeling problem. We introduce a simple procedure to obtain highly competitive summarization performance as based on the ROUGE metric on the CNN-DM corpus with a language modeling loss alone. In addition, we show that the learned model transfers well to new, related tasks, which we demonstrate on the XSum corpus.

%We investigate whether competitive performance can be attained without these approaches, by directly re-purposing  language models without any standard machinery from the neural abstractive summarization toolbox. We introduce a simple procedure to obtain highly competitive summarization performance as based on the ROUGE metric on the CNN-DM corpus with a language modeling loss alone.

\end{abstract}

\section{Introduction}
\label{sec:intro}
Text summarization aims to produce short, coherent natural language summaries of longer-form documents while retaining important information from the original source text. Techniques for this task fall on a point along a continuum between \emph{extractive} and \emph{abstractive} summarization. The former seeks to \emph{extract} grammatically valid subsets of the source document such that, when combined, produce a coherent, shorter text. The latter, as the name suggests, aims to \emph{abstract} away the direct lexical and syntactic choices of the source document, and generate summary text from scratch. 

Neural network approaches to abstractive summarization generally encode the source document into some hidden state or representation, then decode this representation into a summarized, abstracted version of the source document~\cite{rush2015neural,nallapati2016abstractive}. These approaches usually rely on a sequence-to-sequence~\cite{sutskever2014sequence} style architecture, and tend to produce fluent, well formed natural language summaries when coupled with beam search or other decoding techniques.

A major weakness of traditional sequence-to-sequence learning when applied to summarization is the lack of a direct copy mechanism, leading to missing or misrepresented details in decoded summaries~\cite{chopra2016abstractive,nallapati2016abstractive,rush2015neural,zeng2016efficient}. Though attention helps ameliorate this issue by directly learning to focus on specific words or phrases in a source document~\cite{chopra2016abstractive}, many have allowed for an explicit copy mechanism inspired by Pointer Networks~\cite{vinyals2015pointer}, by optimizing a differentiable decision whether to generate new text or directly copy from the source~\cite{gu2016incorporating,zeng2016efficient,see2017get}. %jointly optimizing both a mechanism to generate new text as well as deciding when to copy text directly from the source~\cite{gu2016incorporating,see2017get}. 

Additional components in many neural abstractive summarization systems model semantic coverage~\cite{tu2016modeling,see2017get} and provide guidance on where to attend~\cite{gehrmann2018bottom} in order to directly avoid repetition and ancillary details, while encouraging completeness.

%An additional component in many neural abstractive summarization systems is a notion of modeling semantic coverage~\cite{tu2016modeling,see2017get} and guidance on where to attend~\cite{gehrmann2018bottom} in order to directly avoid repetition and ancillary details while encouraging completeness.

%In addition to directly modeling both the copy-operation and semantic coverage in summarization, r
Recent work has incorporated the use of reinforcement learning to directly optimize objectives of interest that may not be differentiable, but are nonetheless useful for summarization, such as directly optimizing the ROUGE score~\cite{paulus2017deep,li2018actor,celikyilmaz2018deep}.

%At the same time, recent work has shown the benefits of large-scale pretraining on large, unlabeled corpora on a variety of downstream tasks in a transfer learning setting~\cite{peters2018deep,Devlin2018BERTPO,howard2018universal,radford2018improving,radford2019language}. 
%Simultaneously, the benefits of large-scale pretraining on large, unlabeled corpora on a variety of downstream tasks in a transfer learning setting~\cite{peters2018deep,Devlin2018BERTPO,howard2018universal,radford2018improving,radford2019language} has been shown. In particular, it has been shown that large-scale attention-only language modeling via decoder-only transformers~\cite{decoder-transformer} as an unsupervised pretraining task admits the ability to perform zero-shot learning on meaningful tasks involving natural language generation~\cite{radford2019language}.

Simultaneously, \citet{peters2018deep}, \citet{Devlin2018BERTPO}, \citet{howard2018universal}, \citet{radford2018improving}, and \citet{radford2019language}, among others, have shown the benefits of large-scale pretraining on large, unlabeled corpora on a variety of downstream tasks in transfer learning settings. In particular, it has been shown that large-scale, attention-only language modeling via decoder-only transformers~\cite{decoder-transformer} as an unsupervised pretraining task admits the ability to perform zero-shot learning on meaningful tasks involving natural language generation~\cite{radford2019language}.

Motivated by this, we propose a simple method that exhibits competitive performance on abstractive summarization without using sequence-to-sequence architectures or other standard tools in the neural abstractive summarization toolbox, and instead using a decoder-only transformer language model with transfer learning. This further illustrates the utility of finetuning language models trained on open domain text.

% \section{Related Work}
% \label{sec:related-work}

% TODO:

% Work on abstractive summarization

% Work on language modeling with transformers

% Work on using language modeling as an auxiliary task

\section{Model}
\label{sec:model}

% \subsection{Transformer Preliminaries}
\paragraph{Transformer Preliminaries}
Our model builds on previous work utilizing decoder-only Transformers~\cite{decoder-transformer} for jointly learning language modeling and sequence transduction in aligned domains, which limits attention to tokens $0, 1,\ldots,n-1$ for predicting token $n$. Formally, a decoder-only Transformer considers a sequence of one-hot token vectors $T = [t_0, t_1, \ldots, t_{n-1}] \in \{0, 1\}^{V\times n}$, with each $t_i\in\{0, 1\}^V$ where $V$ is the size of the vocabulary. Given an embedding matrix $W_{E}\in\mathbb{R}^{d\times V}$ and a positional encoding matrix $W_P\in\mathbb{R}^{d\times (n-1)}$, the model computes an initial hidden representation $H_0$ as 

\begin{equation}
    \label{eq:transformer-embedding}
    H_{0} = W_{E} T + W_{P} \in \mathbb{R} ^{d \times (n - 1)}
\end{equation}

and computes each subsequent hidden representation as
\vspace*{-.25\baselineskip} 
\begin{equation}
    \label{eq:transformer-hidden}
    H_\ell = \mathsf{TRF}(H_{\ell-1}), \forall \ell\in[1, \ldots, L],
\end{equation}

where $\mathsf{TRF}$ is the transformer block with self-attention, first introduced in \citet{Vaswani2017AttentionIA}. We utilize the modifications provided in~\citet{radford2019language}, such as moving Layer Normalization~\cite{layernorm} to the beginning of each transformer block. The final output is
\vspace*{-.2\baselineskip} 
\begin{equation}
    \label{eq:transformer-output}
    Y = \mathsf{softmax}(W_E^\top H_L) \in [0, 1] ^ {V \times (n-1)}
\end{equation}

% write in the fact that Y = p(...)
where $Y_{i, n-1}$ is the probability assigned to the $n^{\mathrm{th}}$ token being the $i^{\mathrm{th}}$ word in our vocabulary given $t_0,\ldots,t_{n-1}$, and $W_E$ is shared between input and output.

% \subsection{Decoder-only Sequence Transduction for Summarization}
\paragraph{Decoder-only Sequence Transduction for Summarization}

Following \citet{decoder-transformer}, we do not use a sequence-to-sequence approach to sequence transduction, and instead opt to construct a single longer sequence that encodes the full mapping.

Formally, consider a set of paired documents $\mathcal{C} = \{(x, y)\}, \vert\mathcal{C}\vert = N$. For a source-summary pair $(x, y) \in\mathcal{C}$, the source document $x = [x_0, \ldots, x_m]$ and reference summary $y = [y_0, \ldots, y_k]$ are sequences of one-hot token vectors, where we assume $m\gg k$.

To learn this mapping using a language model, we combine $x$ and $y$ using special learnable vectors corresponding to control tokens. In addition, we augment Eq.~\ref{eq:transformer-embedding} to include a \emph{segment-specific} (i.e., source or summary) embedding~\cite{Devlin2018BERTPO}. Finally, we reset the positional encoding for the summary. Our model is fed three sequences (see Eq.~\ref{eq:model-inputs}): a concatenation of the source document and the summary ($S$), positional encodings that reset for the summary component ($P$), and segment-specific encodings for the source and the summary ($Q$). We represent the start of the source document with $\alpha$, the beginning of the summary with $\beta$, and the end of sequence with $\delta$. Additionally, we encode the source segment with $\sigma$ and the summary segment with $\tau$.
% Additionally we construct a vector Q to signal if a token is in soruce or target somain
\vspace*{-.2\baselineskip} 
\begin{equation}
\label{eq:model-inputs}
\begin{aligned}
    S & = [\alpha, x_0, \ldots, x_m, \beta, y_0, \ldots, y_k, \delta] \\
    P & = [0, 1,\ldots, m, m + 1, 0, 1, \ldots, k, k + 1, 0] \\
    Q & = [\sigma, \sigma,\ldots, \sigma, \sigma, \tau, \ldots, \tau, \tau] \\
\end{aligned}
\end{equation}

Thus, our model changes Eq.~\ref{eq:transformer-embedding} by adding the position encoding modification from Eq.~\ref{eq:model-inputs} and an additional trainable weight $W_Q$ representing the segment encoding $Q$, yielding Eq.~\ref{eq:transformer-embedding-ours} while leaving Eq.~\ref{eq:transformer-hidden} and~\ref{eq:transformer-output} unchanged.

\vspace*{-.15\baselineskip} 
\begin{equation}
    \label{eq:transformer-embedding-ours}
    H_{0} = W_{E} S + W_{P} P + W_{Q} Q
\end{equation}

The model is trained via maximum likelihood, where we rewrite $S$ in Eq.~\ref{eq:model-inputs} as $[t_0, t_1, \ldots, t_{m + k + 2}, t_{m + k + 3}]$,
and optimize Eq.~\ref{eq:likelihood} per source-summary pair, where $p(t_i \vert t_0, \ldots, t_{i - 1})$ is obtained from $Y$ in Eq.~\ref{eq:transformer-output}.

\begin{equation}
    \label{eq:likelihood}
    p(S) = \prod_{i=1}^{m + k + 3} p(t_i \vert t_0, \ldots, t_{i - 1})
\end{equation}
\vspace*{-.25\baselineskip} 
\begin{figure}
    \centering
    \includegraphics[width=.49\textwidth]{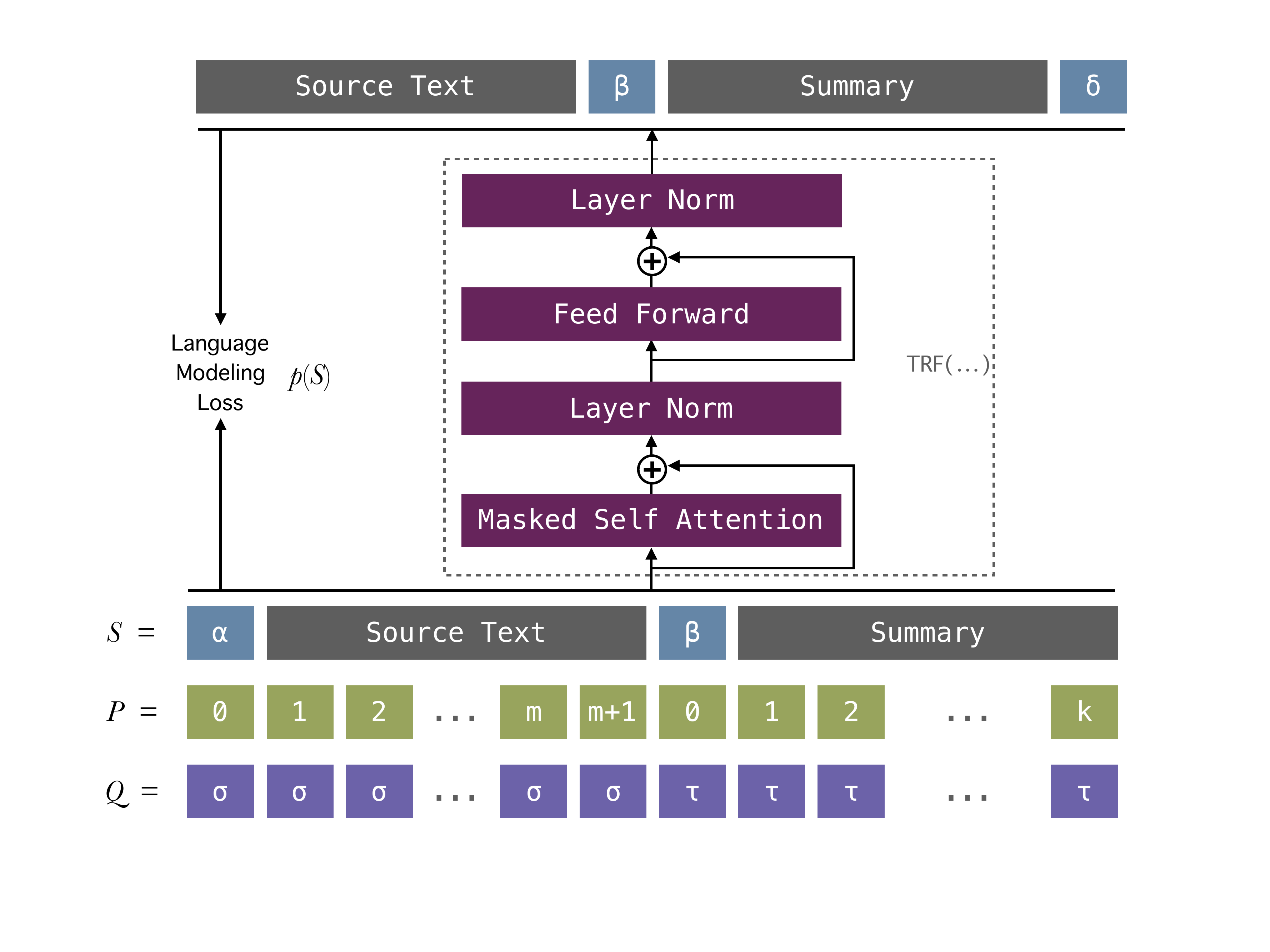}
    \caption{
        Schematic of our Decoder-only Transformer repurposed for summarization. Three input components, $S$, $P$, and $Q$, feed into the masked Transformer (Eq.~\ref{eq:model-inputs} and~\ref{eq:transformer-embedding-ours}). We then optimize the likelihood of the sequence $p(S)$ (Eq.~\ref{eq:likelihood}) which involves predicting the next token, as illustrated by the off-by-one alignment of $S$ on the top and bottom of the figure.
    }
    \label{fig:model}
\end{figure}

% \subsection{Input Representation}
\paragraph{Input Representation}
Given recent trends moving away from purely word- or character-level representations, we utilize data-driven subword encoding via Byte Pair Encoding (BPE)~\cite{sennrich2015neural}, following the procedure outlined in~\citet{radford2019language}. For experiments in which we finetune the 117M parameter model from~\citet{radford2019language}, we utilize their prebuilt vocabulary; in ablation studies, we utilize SentencePiece~\cite{kudo2018sentencepiece} to learn BPE merges.
% \subsection{Learning}

% A natural question that arises out of the representation in Eq.~\ref{eq:model-inputs} is how best to balance predicting the next word for the source document versus predicting the next word for the summary. The naive maximum likelhood objective simply optimizes 

% \begin{equation}
    
% \end{equation}

\section{Experimental Setup}

\paragraph{Datasets} We train and evaluate our models on the CNN/Daily Mail (CNN-DM) corpus~\cite{nallapati2016abstractive} of news articles and summaries\sbox0{\ref{#1}}, utilizing the non-anonymized version~\cite{see2017get}. We use the predefined training, validation, and test splits, and limit source articles to 400 tokens and summaries to 100 tokens at training time.%\footnote{We define token in this context as whitespace/punctuation separated words and punctuation marks. Note that this is not used as the input representation to the model, and is used only as a preprocessing step, as we operate at the subword level.}. 

As an additional test, we train and evaluate the best model configuration from the ablation studies above on the Extreme Summarization (XSum) corpus~\cite{xsum-emnlp}, which contains single sentence summaries of BBC articles. As shown in~\citet{xsum-emnlp}, the XSum corpus requires models to perform a much higher degree of semantic distillation, as indicated by low $n$-gram overlap, high $n$-gram novelty, and poorly performing LEAD-3 baselines.

% \noindent\textbf{Datasets} \qquad We train and evaluate our models on the non-anonymized (as per~\citet{see2017get}) CNN/Daily Mail (CNN-DM) corpus~\cite{nallapati2016abstractive} and the XSum corpus~\cite{xsum-emnlp}. CNN-DM is comprised of news articles and bullet-point summaries\sbox0{\ref{#1}}, 

% For both CNN-DM and XSum, we use the predefined training, validation, and testing splits, and limit source articles to 400 tokens and summaries to 100 tokens at training time.

% In addition, we train and evaluate our best model on the XSum corpus~\cite{xsum-emnlp}, which contains single sentence summaries of BBC articles. As shown in~\citet{xsum-emnlp}, the XSum corpus requires models to perform a much higher degree of semantic distillation, as indicated by $n$-gram overlap as well as LEAD-3 baselines underperforming relative to CNN-DM.

\paragraph{Models \& Inference} In order to illustrate the power and simplicity of this approach, we limit ourselves to minimal hyperparameter tuning. We conduct experiments in two regimes for CNN-DM: first, we \emph{finetune} the model outlined in Sec.~\ref{sec:model} on top of the 117M parameter model release from~\citet{radford2019language}, and second, we perform a full training from scratch in order to ablate the effect of transfer learning. We utilize a context size of 1024 with an embedding dimension of 768, 12 attention heads, and a batch size of 10. We train using the Adam~\cite{kingma2014adam} optimizer with a learning rate of $5\times 10^{-5}$ until the loss ceases to decrease on the validation set. For XSum, we use the highest-performing setup from CNN-DM experiments.

In lieu of beam search, which is commonly used in sequence-to-sequence and transduction models~\cite{sutskever2014sequence,decoder-transformer}, we compare two computationally efficient approaches: greedy decoding, and nucleus sampling~\cite{nucleus-sampling}. In both cases, we decode until we reach the stop-token $\delta$ (Eq.~\ref{eq:model-inputs}). In the case of nucleus sampling, we perform 5 independent decodings\footnote{Nucleus sampling with $p=0.3$ implies we only sample from the top 30\% of of the probability distribution over tokens} with $p=0.3$, then pick the decoding that reports the lowest negative log likelihood score of the \emph{completed summary}, formally represented in Eq.~\ref{eq:criterion}. Note that in Eq.~\ref{eq:criterion} our index begins at $i=m+2$ to account for control tokens, and the fact that we do not wish to account for the likelihood of the source document in our scoring. We use $1/k^{0.6}$ as a normalization term to avoid a preference for shorter summaries, borrowing directly from~\citet{wu2016google}.
\vspace*{-.25\baselineskip} 
\begin{equation}
    \label{eq:criterion}
    -\frac{1}{k ^ {0.6}}\sum_{i=m+2}^{m + k + 3}\log (p(t_i \vert t_0, \ldots, t_m, \ldots, t_{i-1}))
\end{equation}

\begin{table*}[t!]
\small
\centering
% \begin{center}
\tabcolsep=0.11cm
\begin{tabular}{l|c|c|c}
\hline
\bf Method  & \bf ROUGE-1 & \bf ROUGE-2 & \bf ROUGE-L \\ \hline\hline
Pointer-Generator \citep{see2017get}           & 36.44 & 15.66 & 33.42 \\
Pointer-Generator + Coverage \citep{see2017get} & 39.53 & 17.28 & 36.38 \\
ML + RL \citep{paulus2017deep}      & 39.87 & 15.82 & 36.90 \\
% Sentence Rewriting \citep{chen2018fast}  & 40.88 & 17.80 & \textbf{38.54} \\%\hline
Bottom-Up ~\citep{gehrmann2018bottom} & 41.22 & 18.68 & 38.34 \\
DCA (best) ~\citep{celikyilmaz2018deep} & 41.69 & 19.47 & 37.92 \\
GPT-2 \texttt{TL;DR}~\citep{radford2019language} & 29.34 & 8.27 & 26.58 \\
\hline
D-TRF (Finetuned + greedy, \textbf{ours}) & 39.12 & 17.12 & 27.22 \\
D-TRF (Finetuned + nucleus, \textbf{ours}) & 40.70 & 18.03 & 29.62 \\
% D-TRF LM (Scratch + greedy, \bf ours) & WX.YZ & WX.YZ & WX.YZ \\
% D-TRF LM (Scratch + nucleus sampling, \bf ours) & WX.YZ & WX.YZ & WX.YZ \\
\hline
\end{tabular}
% \end{center}
\caption{\label{table:main-results} Comparison of our methods (lower section) with select existing methods on the CNN-DM dataset.}
\end{table*}

\begin{table}[h!]
\centering
\small
% \begin{center}
\begin{tabular}{p{38mm}|p{7mm}|p{7mm}|p{7mm}}
\hline
\bf Ablation  & \bf R-1 & \bf R-2 & \bf R-L \\ \hline\hline
Best & 40.70 & 18.03 & 29.62 \\ \hline
(--) Finetuning & 36.10 & 15.06 & 26.92 \\ 
(--) Segment encoding (Eq. ~\ref{eq:transformer-embedding-ours}) & 38.80 & 16.33 & 27.19  \\ 
\hline
\end{tabular}
% \end{center}
\caption{\label{table:ablation} Ablation of model components on CNN-DM (Decoded via nucleus sampling procedure).}
\end{table}

\begin{table}[t!]
\centering
\small
% \begin{center}
\begin{tabular}{p{42mm}|p{6mm}|p{6mm}|p{6mm}}
\hline
\bf Method  & \bf R-1 & \bf R-2 & \bf R-L \\ \hline\hline
Seq2Seq Baseline & 28.42 & 8.77 & 22.48 \\ 
Conv-Seq2Seq & 31.27 & 11.07 & 25.23 \\ 
Topic-ConvSeq2Seq & 31.89 & 11.54 & 25.75 \\ 
\hline
D-TRF (Finetuned + nucleus) & 34.19 & 12.17 & 27.06\\
\hline
\end{tabular}
% \end{center}
\caption{\label{table:xsum-results} Comparison of our methods (lower section) with select existing methods on XSum, as reported in~\citet{xsum-emnlp}.}
\end{table}

\noindent\textbf{Evaluation} \qquad We evaluate all models using the ROUGE metric~\cite{lin-2004-rouge}, in particular the F1 variants of ROUGE-1, ROUGE-2, and ROUGE-L which measure unigram overlap, bigram overlap, and longest common subsequence respectively. %In addition to evaluating on the CNN-DM dataset on which we train, we also report out-of-domain ROUGE scores on the XSum dataset, which has been shown~\cite{xsum-emnlp} to have desirable properties around lack of lexical overlap with respect to source documents, which encourages models to be more abstractive.

\section{Results}
\label{sec:results}
\paragraph{CNN-DM} Our main results are displayed in Table~\ref{table:main-results}, where we compare our method (in the bottom section of the table) to existing methods (in the upper portion) on the CNN-DM dataset, and show ablations in Table~\ref{table:ablation}.

We note that our models (for ROUGE-1 and -2) are competitive even when using greedy decoding, and without any sequence-to-sequence style architectures or coverage terms, illustrating the power of this approach for abstractive summarization. We note that using a well trained language model~\cite{radford2019language} and then finetuning yields a significant performance jump (as shown via ablation in Table~\ref{table:ablation}), motivating this method in practical contexts given the recent trends toward large-scale, self-supervised learning approaches~\cite{Devlin2018BERTPO,radford2019language,peters2018deep,dai2015semi}.

Our model does not perform well on the ROUGE-L metric, which measures longest-common-subsequence (LCS) between the reference summary and our decoded summary. %We hypothesize that this is due to our models' tendency to alter lexical choices or syntactic structure at decoding time as well as our lack of an explicit coverage mechanism. 
Many (\citet{schluter2017limits} and \citet{lloret2018}, among others) have pointed out deficiencies in ROUGE as an evaluation metric, so we attempt to understand our models deficiencies manually. To investigate, we pick fifty random summaries that score in the bottom 5\% of individual ROUGE-L scores, and examine manually for three traits
\footnote{Examples are included in the Appendix}: fluency, false inclusion (adding extraneous/wrong details), and exclusion (missing details from the reference). We find that 86\% (43/50) of summaries are fluent, 74\% (37/50) exhibited false inclusion, and 92\% (46/50) exhibited exclusion. Of those exhibiting false inclusion, 67\% (31/46) also were marked as exhibiting exclusion. Though not systematic and inconclusive statistically, we believe this indicates that our model suffers from ``distractions'', and attends to details that are not summary worthy as judged by reference summaries. This can systematically limit the highest possible ROUGE-L score our model can achieve due to the fact that LCS requires \emph{interrupted matches}, and skipping over a large subset of the source impairs a models ability to perform well on a metric like ROUGE. Combining our approach with explicitly learned masking methods presented in~\citet{gehrmann2018bottom} may ameliorate these issues by better directing the self-attention mechanism.

%We note that our summaries are quite fluent and on topic, but share a common trait -- we miss key topical details present in the reference summary (for example, the ``abandoned mansion'' in the top row of Table~\ref{table:summary-diagnosis-small}).

\paragraph{XSum}
% \subsubsection{XSum Results}
As a secondary evaluation of our approach, we train our best model on the XSum dataset~\cite{xsum-emnlp} and report ROUGE scores in a direct comparison to the benchmarks reported. Results for these experiments are shown in Table~\ref{table:xsum-results}. We achieve highly competitive performance relative to models reported in~\citet{xsum-emnlp} building on a finetuning approach without using many of the inductive biases traditionally present in summarization methods.

\section{Conclusion}

This work puts forward a simple approach to abstractive summarization by viewing sequence transduction as a language modeling problem. We show the effectiveness of using decoder-only transformers for this task, in particular, when coupled with recent advances in large-scale language modeling and transfer learning. We show that competitive performance on two benchmark datasets is possible without many of the standard tools in neural abstractive summarization, such as sequence-to-sequence modeling, coverage mechanisms, direct ROUGE optimization via reinforcement learning, or beam search, instead relying on a purely language modeling loss and simple decoding mechanisms such as nucleus sampling and greedy decoding. This approach yields highly fluent text, and illustrates the power of unsupervised representation learning-based transfer learning for downstream tasks.

% \section*{Acknowledgments}

% The acknowledgments should go immediately before the references.  Do
% not number the acknowledgments section. Do not include this section
% when submitting your paper for review. \\
% \clearpage
% \newpage
\bibliography{emnlp-ijcnlp-2019}

\begin{thebibliography}{31}
\expandafter\ifx\csname natexlab\endcsname\relax\def\natexlab#1{#1}\fi

\bibitem[{Celikyilmaz et~al.(2018)Celikyilmaz, Bosselut, He, and
  Choi}]{celikyilmaz2018deep}
Asli Celikyilmaz, Antoine Bosselut, Xiaodong He, and Yejin Choi. 2018.
\newblock Deep communicating agents for abstractive summarization.
\newblock \emph{arXiv preprint arXiv:1803.10357}.

\bibitem[{Chopra et~al.(2016)Chopra, Auli, Rush, and
  Harvard}]{chopra2016abstractive}
Sumit Chopra, Michael Auli, Alexander~M Rush, and SEAS Harvard. 2016.
\newblock Abstractive sentence summarization with attentive recurrent neural
  networks.
\newblock \emph{Proceedings of NAACL-HLT16}, pages 93--98.

\bibitem[{Dai and Le(2015)}]{dai2015semi}
Andrew~M Dai and Quoc~V Le. 2015.
\newblock Semi-supervised sequence learning.
\newblock In \emph{Advances in neural information processing systems}, pages
  3079--3087.

\bibitem[{Devlin et~al.(2018)Devlin, Chang, Lee, and
  Toutanova}]{Devlin2018BERTPO}
Jacob Devlin, Ming-Wei Chang, Kenton Lee, and Kristina Toutanova. 2018.
\newblock Bert: Pre-training of deep bidirectional transformers for language
  understanding.
\newblock \emph{CoRR}, abs/1810.04805.

\bibitem[{Gehrmann et~al.(2018)Gehrmann, Deng, and Rush}]{gehrmann2018bottom}
Sebastian Gehrmann, Yuntian Deng, and Alexander Rush. 2018.
\newblock Bottom-up abstractive summarization.
\newblock In \emph{Proceedings of the 2018 Conference on Empirical Methods in
  Natural Language Processing}, pages 4098--4109.

\bibitem[{Gu et~al.(2016)Gu, Lu, Li, and Li}]{gu2016incorporating}
Jiatao Gu, Zhengdong Lu, Hang Li, and Victor~OK Li. 2016.
\newblock Incorporating copying mechanism in sequence-to-sequence learning.
\newblock \emph{arXiv preprint arXiv:1603.06393}.

\bibitem[{{Holtzman} et~al.(2019){Holtzman}, {Buys}, {Forbes}, and
  {Choi}}]{nucleus-sampling}
Ari {Holtzman}, Jan {Buys}, Maxwell {Forbes}, and Yejin {Choi}. 2019.
\newblock \href {http://arxiv.org/abs/1904.09751} {{The Curious Case of Neural
  Text Degeneration}}.
\newblock \emph{arXiv e-prints}, page arXiv:1904.09751.

\bibitem[{Howard and Ruder(2018)}]{howard2018universal}
Jeremy Howard and Sebastian Ruder. 2018.
\newblock Universal language model fine-tuning for text classification.
\newblock In \emph{Proceedings of the 56th Annual Meeting of the Association
  for Computational Linguistics (Volume 1: Long Papers)}, pages 328--339.

\bibitem[{Kingma and Ba(2014)}]{kingma2014adam}
Diederik~P Kingma and Jimmy Ba. 2014.
\newblock Adam: A method for stochastic optimization.
\newblock \emph{arXiv preprint arXiv:1412.6980}.

\bibitem[{Kudo and Richardson(2018)}]{kudo2018sentencepiece}
Taku Kudo and John Richardson. 2018.
\newblock Sentencepiece: A simple and language independent subword tokenizer
  and detokenizer for neural text processing.
\newblock \emph{arXiv preprint arXiv:1808.06226}.

\bibitem[{{Lei Ba} et~al.(2016){Lei Ba}, {Kiros}, and {Hinton}}]{layernorm}
Jimmy {Lei Ba}, Jamie~Ryan {Kiros}, and Geoffrey~E. {Hinton}. 2016.
\newblock \href {http://arxiv.org/abs/1607.06450} {{Layer Normalization}}.
\newblock \emph{arXiv e-prints}, page arXiv:1607.06450.

\bibitem[{Li et~al.(2018)Li, Bing, and Lam}]{li2018actor}
Piji Li, Lidong Bing, and Wai Lam. 2018.
\newblock Actor-critic based training framework for abstractive summarization.
\newblock \emph{arXiv preprint arXiv:1803.11070}.

\bibitem[{Lin(2004)}]{lin-2004-rouge}
Chin-Yew Lin. 2004.
\newblock \href {https://www.aclweb.org/anthology/W04-1013} {{ROUGE}: A package
  for automatic evaluation of summaries}.
\newblock In \emph{Text Summarization Branches Out: Proceedings of the {ACL}-04
  Workshop}, pages 74--81, Barcelona, Spain. Association for Computational
  Linguistics.

\bibitem[{Liu et~al.(2018)Liu, Saleh, Pot, Goodrich, Sepassi, Kaiser, and
  Shazeer}]{decoder-transformer}
Peter~J. Liu, Mohammad~Ahmad Saleh, Etienne Pot, Ben Goodrich, Ryan Sepassi,
  Lukasz Kaiser, and Noam Shazeer. 2018.
\newblock \href {https://openreview.net/pdf?id=Hyg0vbWC-} {Generating wikipedia
  by summarizing long sequences}.

\bibitem[{Lloret et~al.(2018)Lloret, Plaza, and Aker}]{lloret2018}
Elena Lloret, Laura Plaza, and Ahmet Aker. 2018.
\newblock \href {https://doi.org/10.1007/s10579-017-9399-2} {The challenging
  task of summary evaluation: an overview}.
\newblock \emph{Language Resources and Evaluation}, 52(1):101--148.

\bibitem[{Nallapati et~al.(2016)Nallapati, Zhou, Gulcehre, Xiang
  et~al.}]{nallapati2016abstractive}
Ramesh Nallapati, Bowen Zhou, Caglar Gulcehre, Bing Xiang, et~al. 2016.
\newblock Abstractive text summarization using sequence-to-sequence rnns and
  beyond.
\newblock \emph{arXiv preprint arXiv:1602.06023}.

\bibitem[{Narayan et~al.(2018)Narayan, Cohen, and Lapata}]{xsum-emnlp}
Shashi Narayan, Shay~B. Cohen, and Mirella Lapata. 2018.
\newblock Don't give me the details, just the summary! {T}opic-aware
  convolutional neural networks for extreme summarization.
\newblock In \emph{Proceedings of the 2018 Conference on Empirical Methods in
  Natural Language Processing}, Brussels, Belgium.

\bibitem[{Paulus et~al.(2017)Paulus, Xiong, and Socher}]{paulus2017deep}
Romain Paulus, Caiming Xiong, and Richard Socher. 2017.
\newblock A deep reinforced model for abstractive summarization.
\newblock \emph{arXiv preprint arXiv:1705.04304}.

\bibitem[{Peters et~al.(2018)Peters, Neumann, Iyyer, Gardner, Clark, Lee, and
  Zettlemoyer}]{peters2018deep}
Matthew Peters, Mark Neumann, Mohit Iyyer, Matt Gardner, Christopher Clark,
  Kenton Lee, and Luke Zettlemoyer. 2018.
\newblock Deep contextualized word representations.
\newblock In \emph{Proceedings of the 2018 Conference of the North American
  Chapter of the Association for Computational Linguistics: Human Language
  Technologies, Volume 1 (Long Papers)}, pages 2227--2237.

\bibitem[{Radford et~al.()Radford, Narasimhan, Salimans, and
  Sutskever}]{radford2018improving}
Alec Radford, Karthik Narasimhan, Tim Salimans, and Ilya Sutskever.
\newblock Improving language understanding by generative pre-training.

\bibitem[{Radford et~al.(2019)Radford, Wu, Child, Luan, Amodei, and
  Sutskever}]{radford2019language}
Alec Radford, Jeffrey Wu, Rewon Child, David Luan, Dario Amodei, and Ilya
  Sutskever. 2019.
\newblock Language models are unsupervised multitask learners.

\bibitem[{Rush et~al.(2015)Rush, Chopra, and Weston}]{rush2015neural}
Alexander~M Rush, Sumit Chopra, and Jason Weston. 2015.
\newblock A neural attention model for abstractive sentence summarization.
\newblock \emph{arXiv preprint arXiv:1509.00685}.

\bibitem[{Schluter(2017)}]{schluter2017limits}
Natalie Schluter. 2017.
\newblock The limits of automatic summarisation according to rouge.
\newblock In \emph{Proceedings of the 15th Conference of the European Chapter
  of the Association for Computational Linguistics: Volume 2, Short Papers},
  pages 41--45.

\bibitem[{See et~al.(2017)See, Liu, and Manning}]{see2017get}
Abigail See, Peter~J Liu, and Christopher~D Manning. 2017.
\newblock Get to the point: Summarization with pointer-generator networks.
\newblock In \emph{Proceedings of the 55th Annual Meeting of the Association
  for Computational Linguistics (Volume 1: Long Papers)}, pages 1073--1083.

\bibitem[{Sennrich et~al.(2015)Sennrich, Haddow, and
  Birch}]{sennrich2015neural}
Rico Sennrich, Barry Haddow, and Alexandra Birch. 2015.
\newblock Neural machine translation of rare words with subword units.
\newblock \emph{arXiv preprint arXiv:1508.07909}.

\bibitem[{Sutskever et~al.(2014)Sutskever, Vinyals, and
  Le}]{sutskever2014sequence}
Ilya Sutskever, Oriol Vinyals, and Quoc~V Le. 2014.
\newblock Sequence to sequence learning with neural networks.
\newblock In \emph{Advances in neural information processing systems}, pages
  3104--3112.

\bibitem[{Tu et~al.(2016)Tu, Lu, Liu, Liu, and Li}]{tu2016modeling}
Zhaopeng Tu, Zhengdong Lu, Yang Liu, Xiaohua Liu, and Hang Li. 2016.
\newblock Modeling coverage for neural machine translation.
\newblock \emph{arXiv preprint arXiv:1601.04811}.

\bibitem[{Vaswani et~al.(2017)Vaswani, Shazeer, Parmar, Uszkoreit, Jones,
  Gomez, Kaiser, and Polosukhin}]{Vaswani2017AttentionIA}
Ashish Vaswani, Noam Shazeer, Niki Parmar, Jakob Uszkoreit, Llion Jones,
  Aidan~N. Gomez, Lukasz Kaiser, and Illia Polosukhin. 2017.
\newblock Attention is all you need.
\newblock In \emph{NIPS}.

\bibitem[{Vinyals et~al.(2015)Vinyals, Fortunato, and
  Jaitly}]{vinyals2015pointer}
Oriol Vinyals, Meire Fortunato, and Navdeep Jaitly. 2015.
\newblock Pointer networks.
\newblock In \emph{Advances in Neural Information Processing Systems}, pages
  2692--2700.

\bibitem[{Wu et~al.(2016)Wu, Schuster, Chen, Le, Norouzi, Macherey, Krikun,
  Cao, Gao, Macherey et~al.}]{wu2016google}
Yonghui Wu, Mike Schuster, Zhifeng Chen, Quoc~V Le, Mohammad Norouzi, Wolfgang
  Macherey, Maxim Krikun, Yuan Cao, Qin Gao, Klaus Macherey, et~al. 2016.
\newblock Google's neural machine translation system: Bridging the gap between
  human and machine translation.
\newblock \emph{arXiv preprint arXiv:1609.08144}.

\bibitem[{Zeng et~al.(2016)Zeng, Luo, Fidler, and Urtasun}]{zeng2016efficient}
Wenyuan Zeng, Wenjie Luo, Sanja Fidler, and Raquel Urtasun. 2016.
\newblock Efficient summarization with read-again and copy mechanism.
\newblock \emph{arXiv preprint arXiv:1611.03382}.

\end{thebibliography}
\bibliographystyle{acl_natbib}

\clearpage
\appendix

\section{Examples from the manual analysis from Sec.~\ref{sec:results}.}
\begin{table*}[b!]
\centering
\begin{center}
\begin{tabular}{|p{49mm}|p{49mm}|p{49mm}|}
\hline
\textbf{Source} & \textbf{Reference} & \textbf{Ours} \\
\hline
\small[...] mike tyson lived in a southington, ohio, mansion. [...] tyson sold the house in 1999; it's due to become, of all things, a church. the video can be seen at the top of this story [...]  &   \small here are six of cnn's best videos of the week. clips include a look at mike tyson's abandoned mansion.  &   \small former boxer mike tyson sold his house in 1999. it's due to be converted into a church. the video can be seen at the top of this story. \\
\hline
 \small manchester city [...] tests have confirmed the spain international suffered no fractures after being caught in the face [...] pellegrini's men, joint top of the table on new year's day, have slumped to fourth place [...] & \small manchester city defeated west ham 2-0 in their premier league clash. david silva was taken to hospital after a challenge by chiekhou kouyate. spain international has allayed fans' fears with a twitter message. &  \small david silva was injured in the second half of man city's 2-0 win against west ham. the spain international was carried off on a stretcher after eight minutes. manuel pellegrini's side were joint top of the premier league table until their current slump.\\
\hline 
\small at least 15 fortune 500 companies, many of them worth north of a \$1 billion, paid zero income taxes in 2014, says a report out last week from the citizens for tax justice. according to the report, household names like cbs, general electric and mattel all successfully manipulated the u.s. tax code to avoid paying taxes on their massive profits. [...] what's more, most of those 15 were actually given federal tax rebates in the tens or even hundreds of millions. [...] & 
\small many of the companies named in a report out april 9 from the citizens for tax justice even received federal tax rebates. the companies include household names such as cbs, mattel, prudential and time warner. &
\small cbs, general electric and mattel all successfully avoided paying a penny in income taxes. most of those 15 fortune 500 companies managed to get through 2014 without paying a penny in income taxes. \\
\hline
\end{tabular}
\end{center}
\caption{\label{table:summary-diagnosis-small}
Three randomly selected summaries from our model which score in the bottom 5\% of ROUGE-L scores.
}
\end{table*}

Table~\ref{table:summary-diagnosis-small} provides examples of summaries from the bottom 5\% of ROUGE-L scores on CNN-DM for the procedure outlined in Sec.~\ref{sec:results}.

\section{Example Output from CNN-DM}

Table~\ref{table:summary-diagnosis-copy} illustrates the ability of the model to both copy and synthesize.

\begin{table*}[b!]
\centering
\begin{center}
\begin{tabular}{|p{49mm}|p{49mm}|p{49mm}|}
\hline
\textbf{Source} & \textbf{Source} & \textbf{Source} 
\\ \hline
	arsenal, newcastle united and southampton have checked on caen midfielder n'golo kante. paris-born kante is a defensive minded player who has impressed for caen this season and they are willing to sell for around £5million. marseille have been in constant contact with caen over signing the 24-year-old who has similarities with lassana diarra and claude makelele in terms of stature and style. n'golo kante is attracting interest from a host of premier league clubs including arsenal. caen would be willing to sell kante for around £5million.  &   

	wimbledon semi-finalist milos raonic and 19-year-old australian nick kyrgios will make their debuts at the aegon championships at queen's club this summer. canada's raonic, ranked no 6 in the world, lost to roger federer in last year's wimbledon semi-final while kyrgios burst onto the scene with a shock fourth-round victory over two-time champion rafael nadal. the duo will join nadal, andy murray, stan wawrinka, marin cilic and defending champion grigor dimitrov at queen's, which begins on june 15. croatia's milos raonic in action during his run to the wimbledon semi-finals last summer. nick kyrgios was responsible for the biggest upset at sw19 last year when he beat rafael nadal. & 
 
 	jason dufner's marriage has landed in the rough as he agreed a divorce settlement from wife amanda at the end of last month. dufner and wife amanda married in 2012 and were considered one of the golden couples of golf, but the pair separated in february and the divorce was finalised on march 31. according to the divorce settlement, filed on march 16 by amanda, there had been an 'irretrievable breakdown of the marriage' and there was 'a complete incompatibility of temperament that the parties can no longer live together.' jason dufner looks dejected as he struggles on the golf course following a neck injury last year. dufner and amanda during happier times after he won the pga championship in 2012. [...] 
\\ \hline
\textbf{Reference} & \textbf{Reference} & \textbf{Reference} 
\\ \hline
	n'golo kante is wanted by arsenal, newcastle and southampton. marseille are also keen on the £5m rated midfielder. kante has been compared to lassana diarra and claude makelele. click here for the latest premier league news.  & 

	milos raonic, last year's wimbledon semi-finalist, will play at queen's club. australian nick kyrgios will also make his debut in west london. kyrgios knocked rafael nadal out of wimbledon in a huge shock last year. & 

	jason dufner and amanda married in 2012. divorce settlement states there had been an 'irretrievable breakdown of the marriage' amanda will receive \$2.5m as part of the settlement while jason will keep two houses.
\\ \hline
\textbf{Ours} & \textbf{Ours} & \textbf{Ours} 
\\ \hline
	n'golo kante is a defensive minded player who has similarities with lassana diarra and claude makelelele. paris-born kante is attracting interest from a host of premier league clubs. arsenal, newcastle and southampton have checked on the player. & 
	canada's raonic and 19-year-old australian nick kyrgios will make debuts at queen's. canada's raonic lost to roger federer in last year's wimbledon semi-final. kyrgios burst onto the scene with a shock fourth-round victory over two-time champion rafael nadal. &
	jason dufner and wife amanda married in 2012. they were considered one of the golden couples of golf. but the pair separated in february and the divorce was finalised on march 31.

\\ \hline
\end{tabular}
\end{center}
\caption{\label{table:summary-diagnosis-copy}
Three example summaries from our model which illustrate the ability to both copy and synthesize.
}
\end{table*}
\end{document}